\documentclass{article}

\AtBeginDocument{ %fix, anders werkt het niet

}

\makeatletter

\newcommand\Autoref[1]{\@first@ref#1,@}
\def\@throw@dot#1.#2@{#1}% discard everything after the dot
\def\@set@refname#1{%    % set \@refname to autoefname+s using \getrefbykeydefault
    \edef\@tmp{\getrefbykeydefault{#1}{anchor}{}}%
    \def\@refname{\@nameuse{\expandafter\@throw@dot\@tmp.@autorefname}s}%
}
\def\@first@ref#1,#2{%
  \ifx#2@\autoref{#1}\let\@nextref\@gobble% only one ref, revert to normal \autoref
  \else%
    \@set@refname{#1}%  set \@refname to autoref name
    \@refname~\ref{#1}% add autoefname and first reference
    \let\@nextref\@next@ref% push processing to \@next@ref
  \fi%
  \@nextref#2%
}
\def\@next@ref#1,#2{%
   \ifx#2@ and~\ref{#1}\let\@nextref\@gobble% at end: print and+\ref and stop
   \else, \ref{#1}% print  ,+\ref and continue
   \fi%
   \@nextref#2%
}

\makeatother

\usepackage{todonotes}
\usepackage{adjustbox}
\usepackage{forest}
\usepackage{enumitem}
\usepackage{amsmath}
\usepackage[skip=0pt]{caption}

\newcommand{\citeAuthor}[1]{{\hypersetup{citecolor=black}\citeauthor{#1}}}
\newcommand{\citeA}[1]{\citeAuthor{#1} \cite{#1}}

\newcommand{\usericon}[1]{\includegraphics[width=#1]{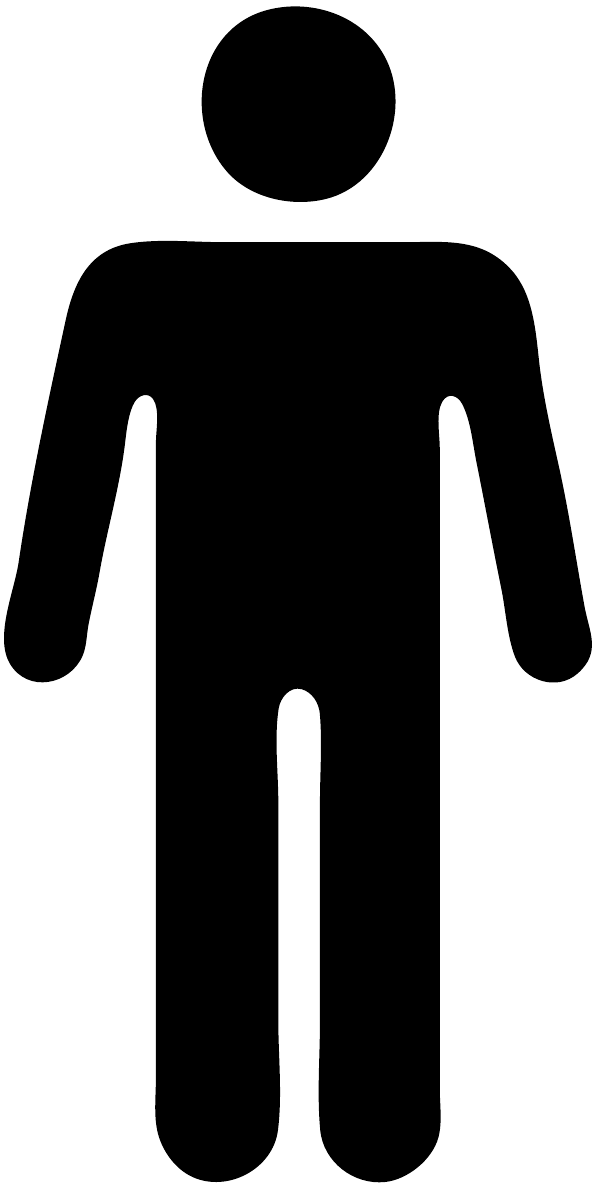}}

\makeatletter
\pgfkeys{/tikz/.cd,
  dogear size/.initial=8pt,
}
\pgfdeclareshape{dogeared}{ % or 'document', or 'file'
  \inheritsavedanchors[from=rectangle]
  \inheritanchorborder[from=rectangle]
  \inheritanchor[from=rectangle]{center}
  \inheritanchor[from=rectangle]{north}
  \inheritanchor[from=rectangle]{south}
  \inheritanchor[from=rectangle]{west}
  \inheritanchor[from=rectangle]{east}

  \savedmacro\dogearsize{%
    \edef\dogearsize{\pgfkeysvalueof{/tikz/dogear size}}%
  }

  \backgroundpath{% this is new
    \southwest \pgf@xa=\pgf@x \pgf@ya=\pgf@y
    \northeast \pgf@xb=\pgf@x \pgf@yb=\pgf@y
    \pgf@xc=\pgf@xb \advance\pgf@xc by-\dogearsize
    \pgf@yc=\pgf@yb \advance\pgf@yc by-\dogearsize
    \pgfpathmoveto{\pgfpoint{\pgf@xa}{\pgf@ya}}
    \pgfpathlineto{\pgfpoint{\pgf@xa}{\pgf@yb}}
    \pgfpathlineto{\pgfpoint{\pgf@xc}{\pgf@yb}}
    \pgfpathlineto{\pgfpoint{\pgf@xb}{\pgf@yc}}
    \pgfpathlineto{\pgfpoint{\pgf@xb}{\pgf@ya}}
    \pgfpathclose
  }
  \foregroundpath{% this is new
    \southwest \pgf@xa=\pgf@x \pgf@ya=\pgf@y
    \northeast \pgf@xb=\pgf@x \pgf@yb=\pgf@y
    \pgf@xc=\pgf@xb \advance\pgf@xc by-\dogearsize
    \pgf@yc=\pgf@yb \advance\pgf@yc by-\dogearsize
    \pgfpathmoveto{\pgfpoint{\pgf@xc}{\pgf@yb}}
    \pgfpathlineto{\pgfpoint{\pgf@xc}{\pgf@yc}}
    \pgfpathlineto{\pgfpoint{\pgf@xb}{\pgf@yc}}
    \pgfpathclose
  }
}
\makeatother

\usetikzlibrary{positioning}
\makeatletter
\tikzset{
    plate/.style={draw=blue, dashed, shape=rectangle, rounded corners=0.5ex, thick, align=right, inner sep=10pt, inner ysep=10pt,label={[text=blue, xshift=-74pt,yshift=14pt]south east:#1}},
    block filldraw/.style={% only the fill and draw styles
        draw, fill=yellow!20},
    block rect/.style={% fill, draw + rectangle (without measurements)
        block filldraw, rectangle},
    block/.style={% fill, draw, rectangle + minimum measurements
        block rect, minimum height=0.8cm, minimum width=6em},
    from/.style args={#1 to #2}{% without transformations
        above right={0cm of #1},% needs positioning library
        /utils/exec=\pgfpointdiff
            {\tikz@scan@one@point\pgfutil@firstofone(#1)\relax}
            {\tikz@scan@one@point\pgfutil@firstofone(#2)\relax},
        minimum width/.expanded=\the\pgf@x,
        minimum height/.expanded=\the\pgf@y}}
\makeatother
\usepackage{microtype}
\usepackage{graphicx}
\usepackage{subfigure}
\usepackage{booktabs} % for professional tables

\usepackage{hyperref}

\usepackage[accepted]{whi2018}

\icmltitlerunning{Instance-Level Explanations for Fraud Detection: A Case Study}

\begin{document}

\twocolumn[
\icmltitle{Instance-Level Explanations for Fraud Detection: A Case Study}



\icmlsetsymbol{equal}{*}

\begin{icmlauthorlist}
\icmlauthor{Dennis Collaris}{tue}
\icmlauthor{Leo M.~Vink}{achmea}
\icmlauthor{Jarke J.~van Wijk}{tue}
% \icmlauthor{Cieua Vvvvv}{goo}
% \icmlauthor{Iaesut Saoeu}{ed}
% \icmlauthor{Fiuea Rrrr}{to}
% \icmlauthor{Tateu H.~Yasehe}{ed,to,goo}
% \icmlauthor{Aaoeu Iasoh}{goo}
% \icmlauthor{Buiui Eueu}{ed}
% \icmlauthor{Aeuia Zzzz}{ed}
% \icmlauthor{Bieea C.~Yyyy}{to,goo}
% \icmlauthor{Teoau Xxxx}{ed}
% \icmlauthor{Eee Pppp}{ed}
\end{icmlauthorlist}

\icmlaffiliation{tue}{Department of Mathematics and Computer Science, Eindhoven University of Technology, The Netherlands}
\icmlaffiliation{achmea}{Achmea BV}
% \icmlaffiliation{to}{Department of Computation, University of Torontoland, Torontoland, Canada}
% \icmlaffiliation{goo}{Googol ShallowMind, New London, Michigan, USA}
% \icmlaffiliation{ed}{School of Computation, University of Edenborrow, Edenborrow, United Kingdom}

\icmlcorrespondingauthor{Dennis Collaris}{d.a.c.collaris@tue.nl}
%\icmlcorrespondingauthor{Eee Pppp}{ep@eden.co.uk}

\icmlkeywords{Fraud Detection, Case Study, Machine Learning, Feature Contribution, Sensitivity Analysis, Local Rule Extraction, Instance-Level Explanations}

\vskip 0.3in
]



\printAffiliationsAndNotice{}

\begin{abstract}

% This document provides a basic paper template and submission guidelines.
% Abstracts must be a single paragraph, ideally between 4--6 sentences long.
% Gross violations will trigger corrections at the camera-ready phase.

Fraud detection is a difficult problem that can benefit from predictive modeling. However, the verification of a prediction is challenging; for a single insurance policy, the model only provides a prediction score.
We present a case study where we reflect on different instance-level model explanation techniques to aid a fraud detection team in their work. To this end, we designed two novel dashboards combining various state-of-the-art explanation techniques. These enable the domain expert to analyze and understand predictions, dramatically speeding up the process of filtering potential fraud cases. Finally, we discuss the lessons learned and outline open research issues.
\end{abstract}

\vspace{-0.5cm}
\section{Introduction}
Many Machine Learning models have been introduced to solve tasks faster and more accurate. However, along with these improvements, the complexity of these models also rapidly increases. This negatively affects the comprehensibility of these models. For instance, Random Forest models are often used for fraud detection. However, for models comprised of hundreds of trees, it can be difficult to grasp which choices are made to yield a prediction. Especially for applications where the consequences of a bad decision are significant and the problem is difficult to predict, an explanation of the choices can be essential for the model to be useful.

To enable model simulatability \cite{Lipton2016}, authors created explanations of the model prediction on a \emph{global} level. However, a simple global explanation may omit many potentially important details, decreasing accuracy with respect to the reference model.
To alleviate this problem, authors have taken a \emph{local} approach: explanations that are simple and remain accurate by only explaining a single instance \cite{Ribeiro2016, lundberg2017unified, robnik2018explanation}.

\newpage

In order to find out how effective these explanations are in a real world application, we conducted a case study at a large insurance firm. To this end, we designed two novel dashboards combining various state-of-the-art explanation techniques, extended where needed. They enable domain experts to analyze and understand individual predictions of Random Forest models. At the insurance firm, the dashboards are used to aid a fraud detection team to more effectively identify potential fraud cases.

The remainder of this paper is structured as follows: we provide an overview of current explanation techniques that are relevant for the interpretation of Random Forest models in \autoref{section:background}. Next, in \autoref{section:casestudy} the case study and dashboards are presented. Applying these techniques in practice revealed many issues and biases that need to be addressed. In \Autoref{section:discussion, section:conclusion}, we reflect on the lessons learned and outline open research issues to stimulate the potential of model explanations.

\section{Related work}\label{section:background}
As Random Forests can get notoriously complex, the interpretability of these models is increasingly important. We can distinguish between two types of approaches. Authors either analyze the features in the context of a model or work on creating a simpler model that behaves and performs like the original model. A visual overview of the taxonomy is shown in \autoref{fig:taxonomy}.

\begin{figure}[H]
\centering

\begin{adjustbox}{valign=t, scale=0.8} % 0.74 when removing the vspace above the feature analysis subsection
\begin{forest}
    for tree={
        draw, 
        rectangle, 
        align=center,
        l sep+=0.2cm,
        s sep+=0.2cm
    }
    [{Random Forest\\explanation}
        [{Feature analysis}
            [{Feature\\importance}]
        	[{Sensitivity\\analysis}]
        	%[{Feature\\interaction}]
        ]
    	[{Model simplification}
    		[{Meta-learning}]
    		[{Model\\condensing}
            	%[{Rule extraction}]
            	%[{Archtype selection}]
            	%[{Decision tree derivation}]
            ]
        ]
    ]
\end{forest}
\end{adjustbox}

\caption{\mbox{\fontsize{8.5pt}{5pt}\selectfont Taxonomy of explanation techniques for Random Forests.}}
\label{fig:taxonomy}
\end{figure}

\vspace{-0.5cm}
\subsection{Feature analysis}
A first approach is to study a feature in isolation and see to what extent it contributes to the predictions made by the model. By understanding which features are more relevant, we reveal information about the decision making process.

\paragraph{Feature importance}
Feature importance metrics enable experts to effectively compare and rank features. They output a single score for a feature based on their contribution to the prediction. 

This can be done globally or locally. In the original implementation of Random Forests by \citeA{breiman2001random}, a global feature importance metric was already included,  which was efficiently estimated due to random subspace projection \cite{ho2002data} in the training process.

Recently there have been efforts to create local feature importance metrics specifically for Random Forests \cite{Palczewska2013, kuz2011interpretation, Altmann2010, tolomei2017interpretable} as well as model-agnostic approaches \cite{lundberg2017unified, vstrumbelj2009explaining, robnik2008explaining}.

\paragraph{Sensitivity analysis}
Another approach to analyze features is through sensitivity analysis \cite{cortez2013using, goldstein2015peeking, Friedman2001, Welling2016, Krause2016a, Lou2013}. This approach analyzes how the output of the model changes when the value of a feature of interest is varied. This is an example of a model-agnostic (or \emph{black box}) approach, as only the input and output of the model are considered. 

\subsection{Model simplification}
Model simplification methods take a reference model and derive a simpler model, while retaining the original behavior as as well as possible. These simplified models are far less complex and thus easier to interpret, but at the expense of generality or accuracy. We distinguish two varieties of methods: \emph{meta-learning}, a black box approach where another model is trained on synthetically generated data from the reference model \cite{Domingos1997, Stiglic2007, Bucilua2006, Zhou2003, Ribeiro2016}, and \emph{model condensing}, which is a white box method that tries to remove the least relevant parts of the model \cite{Assche2007, Perez2007, Gurrutxaga2006, Deng2014, Hara2016}.

\section{Case study: insurance fraud detection}\label{section:casestudy}
\subsection{Problem definition}
A case study is carried out at Achmea BV: one of the leading providers of insurances in the Netherlands. A major concern for this company is fraud. As much as 5\% of insurances are estimated to be fraudulent by the company. However, Achmea is only able to detect a fraction of the estimated amount of fraud. Naturally, there is high interest in automated fraud detection techniques.

\newpage 
 
\begin{figure}[H]
    \centering
    \includegraphics[width=\linewidth]{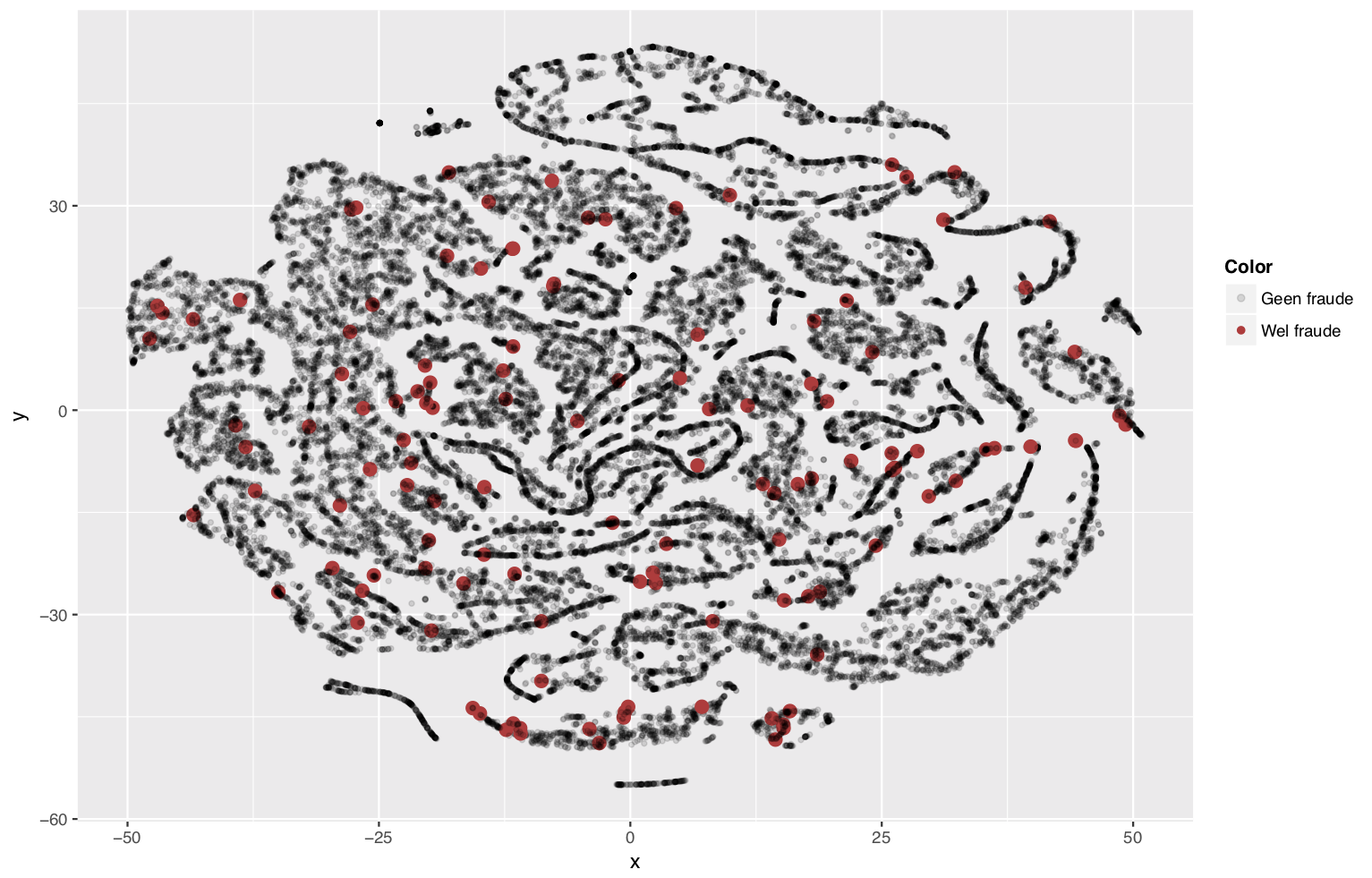}
    \caption{t-SNE projection of sick leave insurances, with 1000 iterations and perplexity 30. Fraud cases are colored red.}
    \label{fig:tsne}
\end{figure}

Fraud detection, however, is a challenging problem. It is fundamentally incomplete \cite{DoshiKim2017Interpretability} in the sense that no perfect rule exists to distinguish a fraudulent case from a non-fraudulent one. To substantiate this, \autoref{fig:tsne} shows a data set of sick leave insurances plotted using t-SNE \cite{maaten2008visualizing}. Similar insurances will appear close to each other in the plot. For any perplexity value, the fraud cases are uniformly distributed among the rest of the data. This clearly shows that no particular subset of the insurances is more likely to contain fraud; fraud seems to appear in all shapes and sizes.

To aid fraud experts in their work, Achmea trained a predictive model to detect fraud among sick leave insurances. A dataset of around 40,000 insurance policies was used, of which 129 records are labeled fraudulent. It contains 49 features sourced from different internal systems; 8 categorical and 41 continuous. To achieve the best possible accuracy, Achmea created a complex bagging ensemble of 100 Random Forests with 500 decision trees each. With an OOB error of 27.7\%, this model still makes mistakes.

The verification of the model prediction is challenging; for a single insurance policy, the fraud expert is only provided with a prediction score (see \autoref{fig:process}). Even if the model is very certain, manual investigation is required to validate the suspicion of fraud.

\begin{figure}[H]
\centering
\begin{adjustbox}{scale=0.75}
\begin{tikzpicture}
\renewcommand\baselinestretch{0.8}
    \node[draw=black, fill=black, text=white, minimum height = 1cm, minimum width = 2cm] (model) {ML Model};
    \node [text width = 1cm, align=center, right = 3cm of model](expert) {\usericon{0.7cm}\\Fraud expert};
    \node [dogeared, thick, draw=red!20!black, minimum height = 1.5cm, align = center, left = 1.5cm of model] (policy) { policy\\data };

    \node [draw, minimum height = 1.0cm, minimum width = 2cm, below = 1.2cm of model] (explainer) {Explainer};

    \path [draw, ->] (model)  -- (expert) node [midway, above] {88\% risk of fraud};
    \path [draw, ->] (policy)  -- (model);
    
    \path [draw, ->] (policy)  |- (explainer.west);
    \path [draw, ->] (model)  -- (explainer);
    \path [draw, ->] (explainer.east)  -| (expert) node [near start, above] {Explanation};
    
     \node[plate=Our contribution, fill opacity=0.07, fill=blue, from={-3.4,-3.1 to 5,-1.4}] {};
\end{tikzpicture}
\end{adjustbox}
\caption{\mbox{\fontsize{8.5pt}{5pt}\selectfont Fraud detection pipeline. Contributions highlighted in blue.}}
\label{fig:process}
\end{figure}

\subsection{\mbox{\fontsize{9.7pt}{5pt}\selectfont Fraud detection augmented with model explanations}}

To provide additional explanations along with the prediction (blue highlights in \autoref{fig:process}), two dashboards were designed. They combine feature importance, sensitivity analysis and model simplification techniques to enable the fraud detection team to more effectively identify potential fraud cases.

\begin{figure}[h]
    \centering
    \begin{subfigure}{\label{fig:feature_dashboard_fi}}\end{subfigure}
    \begin{subfigure}{\label{fig:feature_dashboard_pd}}\end{subfigure}
    \begin{subfigure}{\label{fig:feature_dashboard_histograms}}\end{subfigure}
    \includegraphics[width=\linewidth]{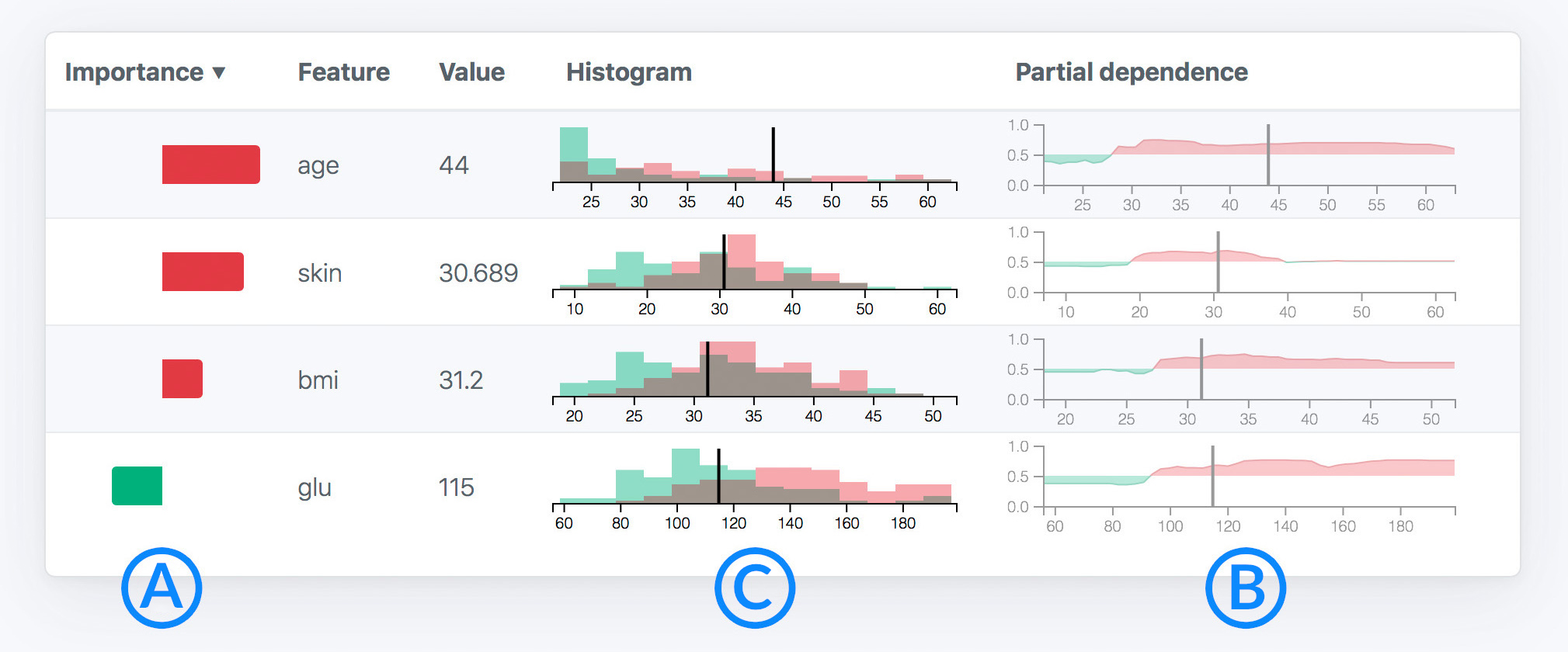}
    \caption{\fontsize{8.5pt}{5pt}\selectfont Feature dashboard for the Pima data set. (A) shows a bar chart expressing feature contribution to the target class. A feature with negative contribution indicates this instance is less likely to belong to the target class. (B) shows partial dependence plots, showing the impact of changing the feature value (indicated with a vertical line) on the final prediction. Along with these model explanations, the distributions of the two classes of training data and the current case are presented (C).}
    \label{fig:feature_dashboard}
\end{figure}

\paragraph{Feature dashboard} This dashboard is centered around features, giving a per-feature explanation of its contribution. The main element is a table of features along with their values for the selected instance, ranked according to feature importance. Various visualization techniques can be chosen and configured, the table can be sorted and tooltips reveal values behind visualizations. To not expose sensitive information, we show an example in \autoref{fig:feature_dashboard} explaining a Random Forest trained on the Pima Indian UCI data set \cite{smith1988using}.

\vspace{-0.35cm}

\begin{figure}[H]
    \centering
    \begin{subfigure}{\label{fig:rule_dashboard_modelmatch}}\end{subfigure}
    \begin{subfigure}{\label{fig:rule_dashboard_vis}}\end{subfigure}
    \includegraphics[width=\linewidth]{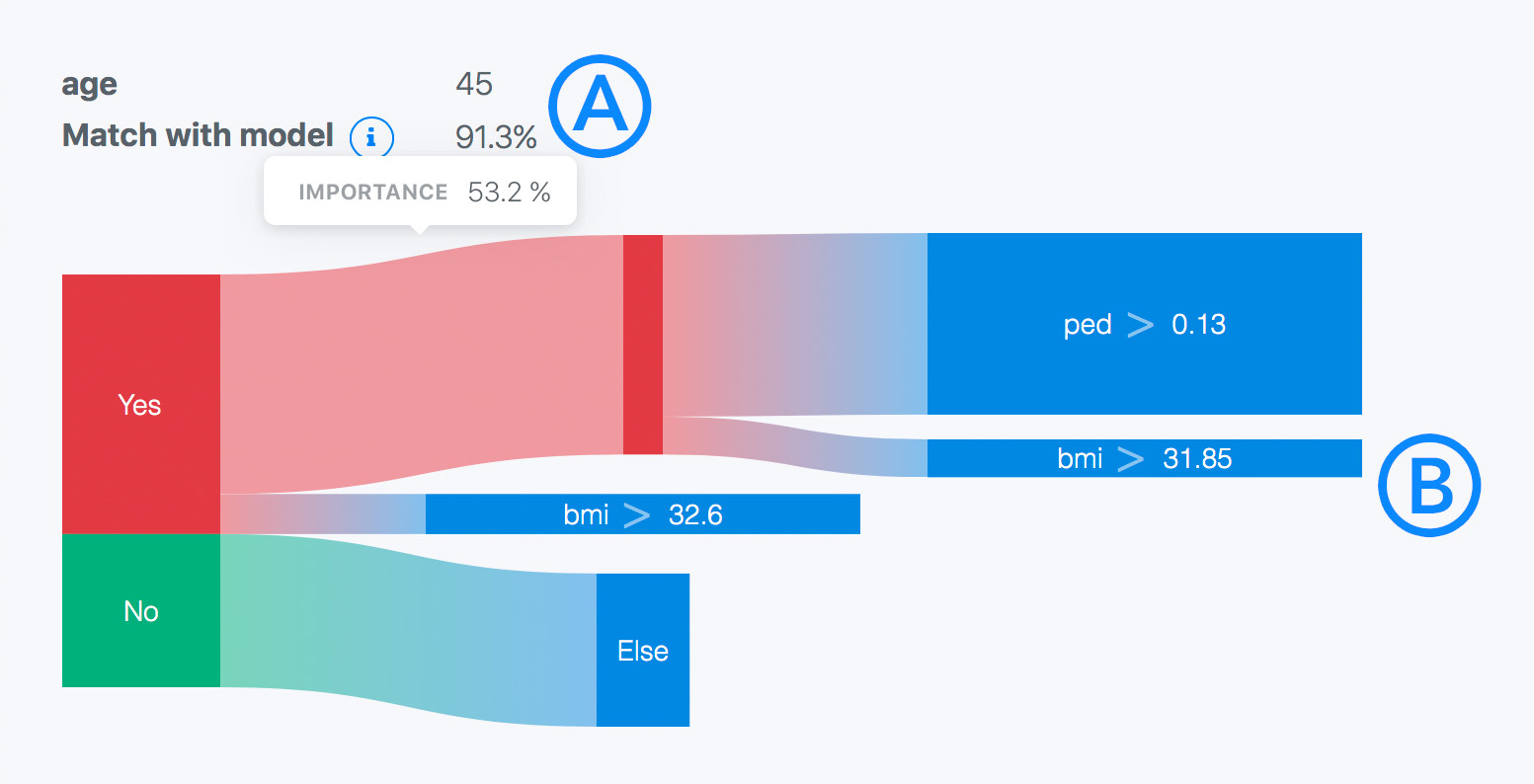}
    \caption{\fontsize{8.5pt}{5pt}\selectfont Rule dashboard for the Pima data set. (B) is a Sankey diagram representation of locally extracted decision rules. As these rules are discarding some detail from the model, an explicit indication of the faithfulness of the model is included (A).}
    \label{fig:rule_dashboard}
\end{figure}

\paragraph{Rule dashboard} The second dashboard takes the possible target classes as a starting point and uses model simplification to present a set of rules that describe the choices the model made for the prediction of those classes. An example is shown in \autoref{fig:rule_dashboard}.

Locally extracted decision rules are visualized as a Sankey diagram. The ratio of color in the first block corresponds to the model posterior probability. Next, a number of rules for the class are connected, where the width of the edge corresponds to the rule importance. Every rule is connected to one or more constraints, where the width of these edges corresponds to the feature contribution.

Clicking on a constraint in the diagram reveals more information about that feature, such as a histogram and partial dependence plot.
As these rules are discarding some details from the model, an explicit indication of the faithfulness of the explanation is included at \autoref{fig:rule_dashboard_modelmatch}.

\subsection{Explanation techniques used}\label{section:techniques}
%To provide additional explanations along with the prediction (blue highlights in \autoref{fig:process}), we used three techniques corresponding with the categories identified in our taxonomy in \autoref{section:background}.
The visualizations in the dashboards are made possible by the following techniques.

\paragraph{Feature contribution}\label{section:featurecontribution} We use the instance-level feature importance method of \citeA{Palczewska2013} for the bar chart in \autoref{fig:feature_dashboard_fi}. It is a white-box approach as it utilizes the structure of the model in order to derive the contribution of a feature to the final prediction. 
It is based on the concept of \emph{local increments}  $LI_f^c$ for a feature $f$ between a parent node $p$ and child node $c$:

\begin{equation}
 LI_f^c = \begin{cases}
    Y_{mean}^c\ -\ Y_{mean}^p, & \text{Parent splits on feature $f$}.\\
    0 & \text{Otherwise}.
  \end{cases}
\end{equation}

where $Y_{mean}^N$ is the probability that an arbitrary element from the training data subset in node $N$ belongs to the target class. This metric is closely related to Gini impurity \cite{breiman1984classification}, the split criterion used for decision trees in a Random Forest, but is specific to the target class.

The feature contribution $FC_{i,t}^f$ for an instance $i$ is first calculated for every tree $t$ in the forest as

\begin{equation}
    FC_{i,t}^f = \sum_{N \in R_{i,t}} LI_f^N
\end{equation}

where $R_{i,t}$ is the composition of all nodes on the path of instance $i$ from the root node to the leaf node in tree $t$. Next, the contribution of a Random Forest can be computed by averaging over all trees.

\paragraph{Partial dependence} 
Feature contribution is unable to capture the influence of the value of a feature on the prediction. To obtain this insight, we use a \emph{sensitivity analysis} technique by \citeA{Friedman2001} called partial dependence. It is visualized using line charts in \autoref{fig:feature_dashboard_pd}. For a local understanding on a feature $f_a$ of instance $i$, $n$ uniformly distributed points are sampled along the range of this feature. Next, $n$ records with values of instance $i$ of all features $f$ with $f \in F, f \neq f_a$ are created and the uniformly sampled values are used for feature $f_a$. Finally, a prediction score is obtained for all $n$ created records and plotted against the values of $f_a$. The resulting curve shows how the prediction score changes when feature $f_a$ in instance $i$ is varied.

For a global understanding, the instance-level sensitivity analysis results for all $k$ training records can be combined. Either the mean of all prediction scores is plotted for each of the $n$ samples \cite{Friedman2001}, or $k$ different lines are plotted to reveal an overall trend \cite{goldstein2015peeking}.

\paragraph{Local rule extraction} By using \emph{model simplification} we can present a simplified model as an explanation. A popular method of doing this is by extracting logical rules. However, to the best of our knowledge, \emph{local} rule extraction techniques have not yet been proposed. To this end, we combined existing approaches to obtain a concise set of decision rules that only has to be faithful locally. These rules are represented as a Sankey diagram in \autoref{fig:rule_dashboard_vis}.

First, a synthetic pruning data set is obtained in the local vicinity of instance $i$ of interest. This can be done by uniformly sampling from an $n$-ball with $n = |F|$, radius $r = \delta$ and centered at instance $i$. These records are labeled by the reference Random Forest. This method is similar to the method used by \citeA{Ribeiro2016} but yields discrete records rather than weighted ones.

Next, all decision rules applicable to instance $i$ are extracted from the Random Forest by extracting the path from root to leaf node when classifying instance $i$ for every tree. These decision rules are first pruned by iteratively removing constraints from the rule and leaving them out when the impact on the prediction on the synthetic pruning set is not worse than a given threshold. Duplicates introduced as result of pruning are removed.

Finally, the relevance of each rule is estimated by a technique introduced by \citeA{Deng2014}. A binary matrix is created with the synthetic pruning data along the rows and set of rules along the columns. Another Random Forest is trained to predict the labels of the synthetic pruning data. The global feature importance from this Random Forest now constitutes a metric of importance for individual rules. By using regularization \cite{deng2013gene}, this importance metric can be biased to favor shorter rules. Discarding irrelevant rules with rule importance below a given threshold yields a set of relevant rules that are locally relevant around instance $i$.

\section{Discussion}\label{section:discussion}
We have applied our methods to the sick leave insurance data and presented the results to five fraud experts. In general, they were very positive and considered it as a highly useful tool to accelerate their understanding. From the different dashboards, they preferred the rule-based version, as they found it to be clear and concise. The partial dependence plots were less appreciated, but for their cases, most of these showed almost flat curves. However, this application in practice also revealed many issues and biases that need to be addressed.

% understanding
\paragraph{Understanding explanations} First and foremost, it was challenging to evaluate explanations. Even though recent literature tries to address this issue \cite{Lipton2016, DoshiKim2017Interpretability}, the community is far from reaching consensus on what best practices are.

We applied three explanation techniques that yielded different results for the insurance case; features with the highest \emph{contribution} did not correspond with features with the highest variance in \emph{partial dependence}. Likewise, the most important \emph{local rules} used yet another set of important features. These explanations may be equally valid and useful, but do not establish trust in the system, nor will they provide a coherent explanation when combined. More research is needed to understand the solution space of possible explanations and to identify trade-offs between desiderata.

Alarmingly, this incongruency did not affect the evaluation by both fraud team nor various data science teams at the insurance firm. They readily trusted the provided explanation and did not question their validity, even when provoked. There seems to be an Illusion Of Explanatory Depth \cite{keil2006explanation} causing overconfidence of understanding and the disregard of uncertainties.
This can be especially dangerous considering various works on the topic of explainability evaluate their systems by means of user testing \cite{DoshiKim2017Interpretability, Ribeiro2016, tolomei2017interpretable}.

Another issue is that the fraud experts confused the presented explanations for actual causality. Using explanations in this context only provides a conjecture on what possible causalities may exist, based on the correlations found by the classifier. We should be very careful not to present misleading explanations to our users.

% data
\paragraph{Data quality} The real world data set introduced more difficulties as compared to standardized UCI data sets. Missing values and imbalanced data have a significant impact on the interpretability, and should always be considered when creating explanations. For instance, if the feature cannot be meaningfully explained, this will have a direct impact on the interpretability of the explanation, regardless of the classifier. 

Likewise, the value of a feature can also lose meaning by imputed values that do not follow the feature distribution (e.g., 9999). In our project imputations skewed histograms obscuring the actual trend, and shifted the decision boundary for features to unrealistic values (e.g., the constraint \emph{Fraud} when the duration of sickness is less than 50 years would only select non-imputed values). 

Additionally, heavy imbalance can make showing histograms of data impossible without some form of normalization. This, in turn, can mislead the expert.

% global local
\paragraph{Generality} We found that global insights are often too simplistic to capture the behavior of a complex model. The fraud model has various different 'strategies' to detect fraud, that will be lost when averaging over all local effects like done with feature importance metrics and partial dependence. 

The latter technique did not even work in a local setting for the complex model: no single feature had a significant impact on the prediction. Rather, the prediction is based on various features in unison. Such interactions are not captured in partial dependence.

The Random Forest model tested on was vastly complex (1.3 million decisions), but we were still able to obtain simple and reasonably accurate explanations by considering single instances. However, even though instance-level explanations offer a solution for this case, we argue this makes it challenging to explore what is happening on the global level; exploring many instance-level explanations is impractical and inefficient for this purpose. Additionally, local explanations may again be misleading, as the expert may falsely assume that the presented behavior applies in more situations than just the instance.

\section{Conclusion}\label{section:conclusion}
In order to find out how effective these explanations are in a real world application, we have conducted a case study at Achmea. We created two dashboards to enable domain expert to analyze and understand individual predictions. The local focus allowed us to explain a very complex model, but at the cost of generality of the explanation. We found that different explanation techniques may yield widely varying results, yet may all be considered reasonably valid and useful. This incongruency is unclear to the domain experts, who were eager to trust any explanation provided to them. This can be especially dangerous for application grounded evaluation of explanation techniques. Finally, data quality can have a significant impact on the explanation and should not be taken for granted.

\newpage

% there is an opportunity to 
%\bibliography{bibliography.bib}

\bibliographystyle{icml2018}

\end{document}